\documentclass{article}
\usepackage{iclr2027_conference,times}
\usepackage[hyphens]{url}
\usepackage{graphicx}
\usepackage{caption}
\usepackage{subcaption}
\usepackage{booktabs}
\usepackage{needspace}
\usepackage{algorithm}
\usepackage{algorithmic}
\usepackage{amsmath,amssymb,amsthm}
\usepackage{xcolor}
\usepackage{hyperref}
\hypersetup{hidelinks,pdfkeywords={Graph Anomaly Detection; Adapt-Then-Combine; Node-Context Disagreement; Diffusion Models}}
\usepackage{tikz}
\usetikzlibrary{shapes.geometric, arrows.meta, positioning, calc, backgrounds, fit, patterns}
\makeatletter
\renewcommand\section{\@startsection{section}{1}{\z@}%
  {-1.55ex plus -0.3ex minus -0.15ex}%
  {1.0ex plus 0.15ex minus 0.1ex}%
  {\large\sc\raggedright}}
\renewcommand\subsection{\@startsection{subsection}{2}{\z@}%
  {-0.75ex plus -0.15ex minus -0.08ex}%
  {0.20ex plus 0.05ex}%
  {\normalsize\sc\raggedright}}
\renewcommand\subsubsection{\@startsection{subsubsection}{3}{\z@}%
  {-0.65ex plus -0.12ex minus -0.06ex}%
  {0.15ex plus 0.04ex}%
  {\normalsize\sc\raggedright}}
\renewcommand\paragraph{\@startsection{paragraph}{4}{\z@}%
  {0.45ex plus 0.1ex minus 0.05ex}%
  {-1em}%
  {\normalsize\bf}}
\makeatother
\newtheorem{prop}{Proposition}
\newtheorem{theorem}{Theorem}
\newtheorem{corollary}{Corollary}

\newenvironment{Proof}{\par\noindent\textbf{Proof.}\nobreakspace\ignorespaces}{\unskip\nobreak\ensuremath{\square}\par}

\newcommand{\gG}{\mathcal{G}}
\newcommand{\sV}{\mathcal{V}}
\newcommand{\sE}{\mathcal{E}}
\newcommand{\sA}{\mathcal{A}}
\newcommand{\sN}{\mathcal{N}}
\newcommand{\sM}{\mathcal{M}}
\newcommand{\R}{\mathbb{R}}
\newcommand{\E}{\mathbb{E}}

\newcommand{\mI}{\mathbf{I}}
\newcommand{\mA}{\mathbf{A}}
\newcommand{\mX}{\mathbf{X}}
\newcommand{\mW}{\mathbf{W}}
\newcommand{\mH}{\mathbf{H}}
\newcommand{\mC}{\mathbf{C}}
\newcommand{\mR}{\mathbf{R}}
\newcommand{\mB}{\mathbf{B}}
\newcommand{\mJ}{\mathbf{J}}
\newcommand{\mPi}{\boldsymbol{\Pi}}
\newcommand{\mU}{\mathbf{U}}
\newcommand{\mZ}{\mathbf{Z}}
\newcommand{\mS}{\mathbf{S}}
\newcommand{\vx}{\mathbf{x}}
\newcommand{\vy}{\mathbf{y}}
\newcommand{\vh}{\mathbf{h}}
\newcommand{\vz}{\mathbf{z}}

\newcommand{\ve}{\mathbf{e}}
\newcommand{\vu}{\mathbf{u}}

\graphicspath{{Figures/}}

\DeclareCaptionStyle{ruled}{labelfont=normalfont,labelsep=colon,strut=off}

\title{\raggedright\fontsize{13.2}{15.8}\selectfont
DDGAD: Disagreement-Driven Graph Anomaly Detection\\
via Adapt-Then-Combine}

\author{
  Yuxin Yang$^1$, Limei Hu$^1$, Feng Chen$^{1,}$\thanks{Corresponding author.} \\
  $^1$College of Artificial Intelligence, Southwest University, Chongqing, China \\
  {\small\texttt{tianyiyoung127@gmail.com}, \ \texttt{hlm0903@swu.edu.cn}, \ \texttt{fengchen.uestc@gmail.com}}
}

\iclrfinalcopy

\begin{document}

\maketitle

\begin{abstract}
Graph anomaly detection (GAD) commonly relies on message passing to jointly encode node attributes and neighborhood context. However, once the two are mixed, an abnormal post-encoding state may reflect either an intrinsic node deviation or incompatible contextual influence, making its source ambiguous. We propose Disagreement-Driven Graph Anomaly Detection (DDGAD), which treats persistent incompatibility between node-wise and contextual estimates as anomaly evidence. Inspired by Adapt-Then-Combine (ATC), DDGAD reverses its consensus objective: Adapt produces a node-wise estimate without new same-step neighbor aggregation, while Combine forms a neighborhood-dependent contextual estimate, and their pre-consensus disagreement is accumulated across ATC steps for detection. We further characterize this signal from graph-spectral and source-response perspectives and derive sufficient conditions for anomaly--normal separation under contextual mixing. Experiments on six benchmarks show the highest average AUROC among the evaluated methods, while controlled interventions and Adapt-operator controls further support persistent disagreement as an effective detection signal.

\vspace{0.8ex}
\end{abstract}

\section{Introduction}
\label{sec:introduction}
Many GNN-based graph anomaly detection (GAD) methods use message passing to jointly encode node attributes and neighborhood context~\citep{ding2019deep,fan2020anomalydae,liu2021anomaly}. This joint encoding makes anomaly detection inherently context-dependent: the representation of a node reflects not only its own attributes but also evidence aggregated from its neighborhood. When the two are incompatible, however, the resulting representation alone does not reveal whether the node state is supported or contradicted by its local context. This creates a \emph{node--context ambiguity} in anomaly scoring.

Existing GAD methods exploit reconstruction discrepancies~\citep{ding2019deep,fan2020anomalydae,li2025diffgad} or neighborhood-consistency signals~\citep{liu2021anomaly,hou2022graphmae,xu2022conad}. However, these signals are typically evaluated on graph-conditioned representations in which node and neighborhood information have already interacted. As a result, node--context compatibility remains implicit in the final detection score. A more direct approach is to construct distinct post-encoding estimates from the same graph-conditioned state---one updated node-wise and the other formed from neighborhood consensus---and explicitly measure whether the two remain compatible.

Our key observation is that compatibility between these two estimates is itself informative. If a node-wise estimate remains compatible with the consensus implied by its neighbors, the two estimates should remain close; persistent disagreement instead indicates a local inconsistency that is difficult to reconcile. We therefore introduce \emph{\mbox{Disagreement-Driven Graph Anomaly Detection}} (DDGAD), which preserves and accumulates node--context disagreement as anomaly evidence.

\begin{figure}[H]
  \centering
  \includegraphics[width=0.85\textwidth]{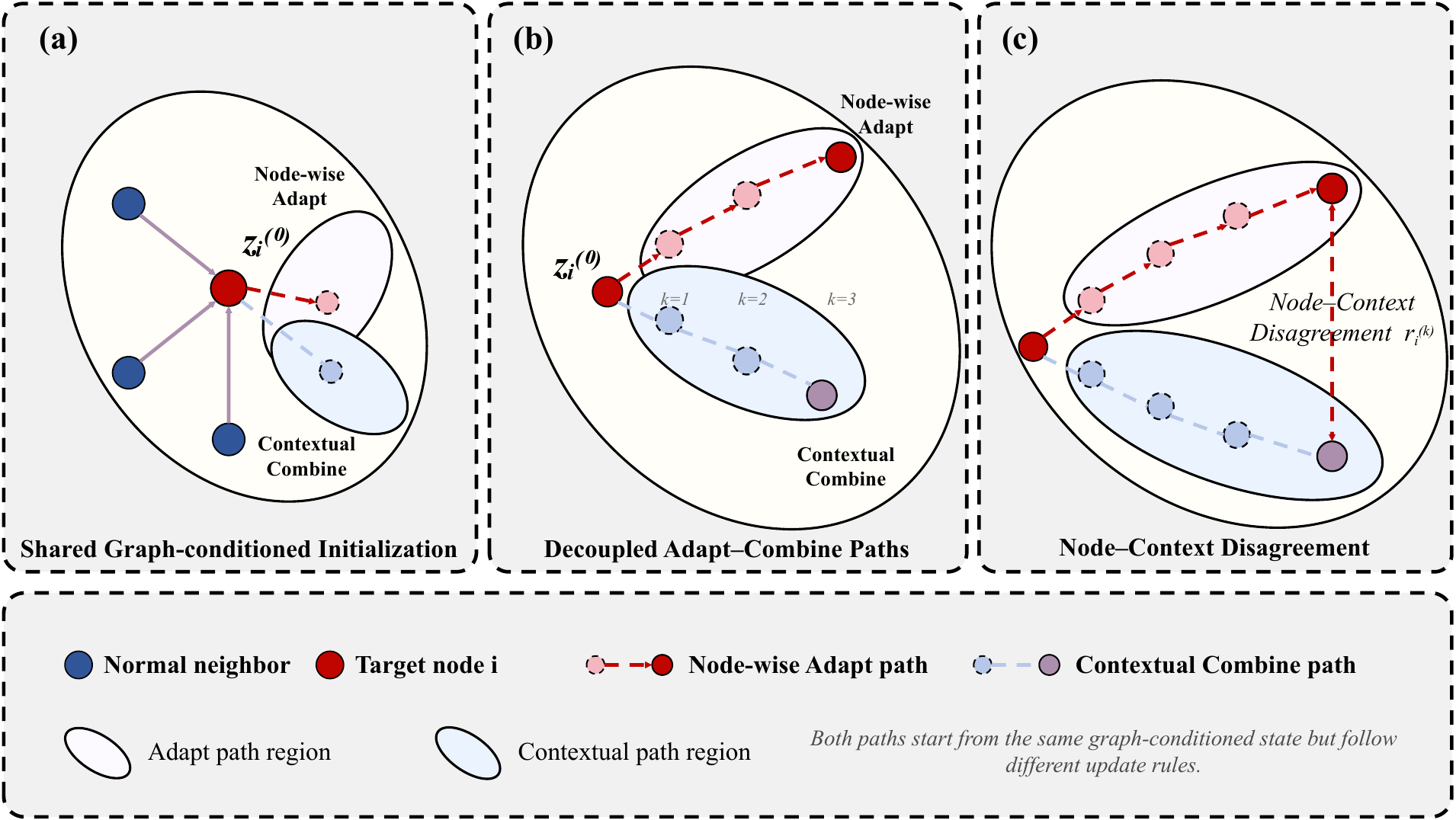}
  \caption{Mechanism illustration of node--context disagreement in DDGAD. Given a shared graph-conditioned state, node-wise Adapt and neighborhood-dependent Combine form estimates under different update rules. Their separation schematically represents node--context incompatibility, with persistent disagreement serving as anomaly evidence.}
  \label{fig:framework}
\end{figure}

\begin{figure*}[t]
  \centering
  \includegraphics[width=0.95\textwidth]{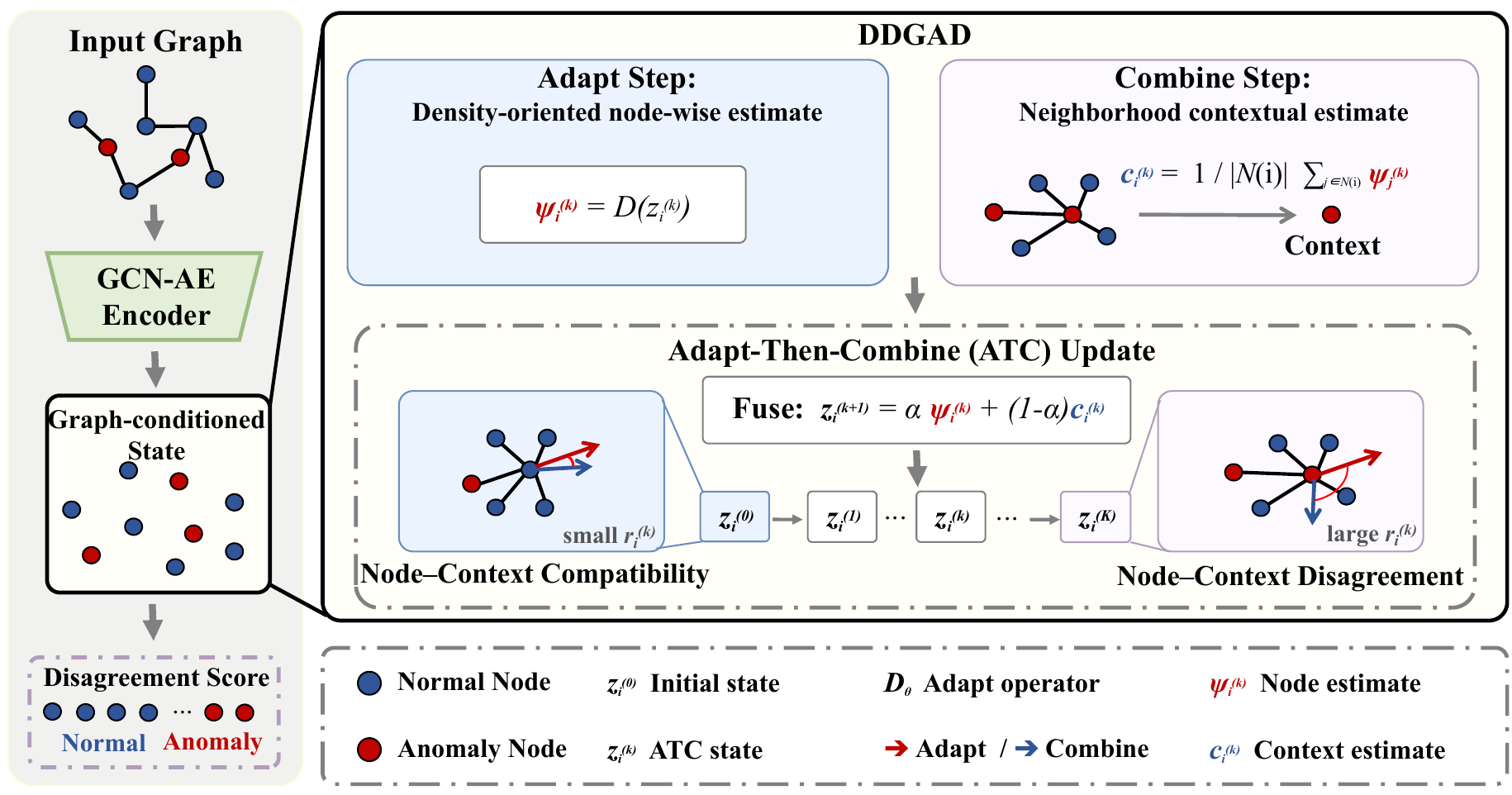}
  \caption{Overall framework of DDGAD. A GCN autoencoder provides the graph-conditioned initialization $\vz_i^{(0)}$. At each ATC step, node-wise Adapt produces $\psi_i^{(k)}$, uniform Combine constructs the contextual estimate $c_i^{(k)}$, and Fuse updates $\vz_i^{(k+1)}$. Their pre-consensus disagreement is accumulated over $K$ steps to obtain the anomaly score.}
  \label{fig:pipeline}
\end{figure*}

Inspired by the Adapt-Then-Combine (ATC) estimation paradigm~\citep{sayed2014adaptation}, DDGAD repurposes consensus formation for anomaly detection. To make node--context ambiguity explicit, DDGAD starts from a shared graph-conditioned state and constructs two post-encoding estimates under different update rules. A diffusion-based Adapt operator~\citep{karras2022elucidating} forms a density-oriented node-wise estimate without new same-step neighbor aggregation, while uniform Combine forms a neighborhood-dependent contextual estimate from neighboring adapted states. Their discrepancy directly reflects node--context compatibility, while persistent disagreement across ATC steps is accumulated for detection. Figure~\ref{fig:framework} illustrates this mechanism conceptually, whereas Figure~\ref{fig:pipeline} presents the full iterative DDGAD framework.

Our contributions are summarized as follows:
\begin{itemize}
    \item We introduce a disagreement-driven perspective for GAD that starts from a graph-conditioned initialization and, at each ATC step, contrasts a node-wise Adapt estimate with a neighborhood-dependent contextual Combine estimate. This makes node--context incompatibility directly observable, and persistent pre-consensus disagreement becomes anomaly evidence.
    \item We characterize the rejected consensus energy spectrally, derive a perturbed population-separation condition from neighborhood composition, and obtain conditional forcing dynamics, a degree-controlled source-response law, and a high-probability persistence bound.
    \item Experiments on six benchmarks establish strong average AUROC; simple Adapt controls support the generality of disagreement-driven scoring, while matched post-encoder interventions on two Amazon graphs isolate a source-specific response consistent with the theoretical analysis, beyond the effect of added degree.
\end{itemize}

\section{Related Work}

Graph anomaly detection spans classical and deep approaches; broader taxonomies, resources, and open challenges are reviewed by Akoglu et al.~\citep{akoglu2015graph} and Ma et al.~\citep{ma2023survey}.

\subsection{Shallow and Reconstruction-based GAD}
Classical non-neural detectors include LOF~\citep{breunig2000lof}, Isolation Forest~\citep{liu2008isolation}, structural clustering with SCAN~\citep{xu2007scan}, and residual-based Radar~\citep{li2017radar}. Deep reconstruction methods score attribute or structural recovery errors, including DOMINANT~\citep{ding2019deep}, AdONE and DONE~\citep{bandyopadhyay2020outlier}, AnomalyDAE~\citep{fan2020anomalydae}, and diffusion-based DiffGAD~\citep{li2025diffgad}. Although their latent models differ, these methods do not explicitly compare a node-wise estimate with a separately constructed neighborhood estimate.

\subsection{Neighborhood consistency-based GAD}
Contextual detectors exploit neighborhood consistency through contrastive or self-supervised objectives, as in CoLA~\citep{liu2021anomaly}, CONAD~\citep{xu2022conad}, and GraphMAE~\citep{hou2022graphmae}. Their compared representations remain neighborhood-dependent, so unreliable context can affect both representation formation and anomaly scoring. DDGAD instead separates post-encoding node-wise and contextual computations before measuring disagreement.

Methods that explicitly model contextual reliability are particularly related. DualGAD captures feature--structure inconsistency through generative and cluster-guided contrastive objectives~\citep{tang2024dualgad}, DC-AES suppresses unreliable propagation using abnormal-edge masking~\citep{li2025dcaes}, and GRAM attributes anomaly decisions through gradient attention maps~\citep{yang2024gram}. DDGAD instead constructs a contextual counter-estimate and retains rejected pre-consensus disagreement as anomaly evidence.

\subsection{Diffusion and Adapt-Then-Combine}
Diffusion has been used for counterfactual anomaly generation in CAGAD~\citep{xiao2024cagad}, graph enhancement in DEGAD~\citep{pang2024degad}, and latent reconstruction in DiffGAD~\citep{li2025diffgad}. DDGAD instead uses EDM as a density-oriented realization of node-wise Adapt, while detection arises from disagreement with a separately computed contextual estimate.

ATC adapts each agent locally and then combines neighboring estimates toward consensus~\citep{sayed2014adaptation}. DDGAD preserves this decomposition but reverses its task semantics: Combine provides a contextual counter-estimate, and the pre-consensus error is retained for detection rather than suppressed. Denoising supplies one realization of node-wise Adapt through the score of a smoothed density~\citep{vincent2011connection,alain2014regularized,stanczuk2022diffusion}.

\section{Methodology}

The transfer from ATC is semantic rather than modular: the Adapt--Combine decomposition remains recognizable, but the pre-consensus quantity that classical ATC suppresses becomes the detector output. We first formalize the ambiguity created by joint encoding and then define two distinct post-encoding estimates from the same graph-conditioned state.

\subsection{Problem formulation}
\label{sec:problem}
Given an undirected attributed graph $\gG=(\sV,\sE)$ with adjacency matrix $\mA\in\{0,1\}^{N\times N}$ and node features $\vx_i\in\R^{d_x}$ stacked into $\mX\in\R^{N\times d_x}$, we rank a small anomalous subset $\sA\subset\sV$ without labels in training or score construction. DDGAD targets contextual anomalies and intrinsic deviations that induce node--context incompatibility; purely intrinsic anomalies that remain context-compatible are outside its separation guarantee. We use $d_x$ for input-feature dimension and $d_z$ for AE latent dimension throughout.

\subsection{Motivation: Ambiguity after Joint Encoding}
\label{sec:motivation}
Message passing jointly encodes node attributes and neighborhood context:
\begin{equation}
\vz_i = f_\theta(\vx_i, \{\vx_j : j\in\sN(i)\}),
\label{eq:joint-encoding}
\end{equation}
so that a graph-conditioned embedding contains both node and contextual contributions. In a linearized $L$-layer GCN with normalized adjacency $\widetilde{\mA}$ and composed feature map $\mW_{1:L}$, a perturbation $\Delta\mX$ on anomalous nodes gives
\begin{equation}
\Delta\mH^{(L)}=\widetilde{\mA}^{L}\Delta\mX\mW_{1:L},\qquad
\Delta\vh_i^{(L)}=\sum_{j\in\sA}(\widetilde{\mA}^{L})_{ij}\Delta\vx_j\mW_{1:L}.
\label{eq:gcn-contamination}
\end{equation}
Hence contextual influence can propagate along length-$L$ walks and alter a node's graph-conditioned state. A single mixed nonlinear embedding does not explicitly expose whether a node-wise estimate remains compatible with the consensus implied by its neighborhood. DDGAD therefore constructs two distinct post-encoding estimates at each ATC step: a node-wise Adapt estimate and a neighborhood-dependent contextual Combine estimate. Their disagreement directly exposes node--context incompatibility; Appendix~\ref{app:theory} gives the propagation argument.

\subsection{Consensus-to-Detection through Inverted ATC}
\label{sec:atc-framework}
The transfer from ATC is semantic rather than modular: DDGAD preserves the Adapt--Combine decomposition but repurposes it for detection. Starting from the graph-conditioned initialization $\vz_i^{(0)}$, DDGAD iterates a shared ATC state $\vz_i^{(k)}$. At step $k$, Adapt produces the node-wise estimate $\psi_i^{(k)}$ without new same-step neighbor aggregation, while Combine forms the neighborhood-dependent contextual estimate $c_i^{(k)}$ from neighboring adapted states. Rather than suppressing their discrepancy before Fuse, DDGAD retains and accumulates it as anomaly evidence.

\paragraph{Adapt.}
Adapt constructs a node-wise reference
\begin{equation}
\psi_i^{(k)} = \mathcal{D}_\theta(\vz_i^{(k)};\sigma_k,\boldsymbol{\xi}_i^{(k)}),
\label{eq:adapt}
\end{equation}
using a shared MLP-based EDM denoiser, noise level $\sigma_k$, and sampling noise $\boldsymbol{\xi}_i^{(k)}$. Conditioned on the current states and noises, Adapt reads neither graph topology nor $\vz_j^{(k)}$ for $j\neq i$, so it performs no new same-step neighbor aggregation. For $k\geq1$, however, $\vz_i^{(k)}$ may already contain cross-node information introduced by the preceding Fuse. Appendix~\ref{app:theory} gives the conditional Jacobian statement and the mixture-aware density interpretation.

\paragraph{Combine.}
Combine uses the uniform mean of adapted neighbors:
\begin{equation}
c_i^{(k)}=
\begin{cases}
\displaystyle\frac{1}{|\sN(i)|}\sum_{j\in\sN(i)}\psi_j^{(k)}, & |\sN(i)|>0,\\[5pt]
\psi_i^{(k)}, & |\sN(i)|=0.
\end{cases}
\label{eq:combine}
\end{equation}
This parameter-free operator deliberately preserves the ordinary neighborhood context. Its difference from Adapt provides the node--context incompatibility signal used for detection.

\paragraph{Fuse.}
With self-confidence $\alpha\in[0,1]$, Fuse updates
\begin{equation}
\vz_i^{(k+1)} = \alpha\,\psi_i^{(k)} + (1-\alpha)\,c_i^{(k)}.
\label{eq:fuse}
\end{equation}
This convex update retains ATC's distributed-estimation interpretation; its local objective is given in Appendix~\ref{app:theory}. Uniform Combine preserves the observed neighborhood context as the contextual estimate against which Adapt is compared.

\subsection{From Consensus Error to Detection Evidence}
Classical ATC regards $(\mI-\mW)\boldsymbol{\Psi}^{(k)}$ as a transient error to be dissipated. DDGAD exposes the same quantity before Fuse and accumulates its node-wise magnitude. Here $\mW$ is the fixed row-normalized neighborhood matrix (with an identity row for an isolated node), and $\boldsymbol{\Psi}^{(k)}$ stacks adapted states. The defining principle is the reinterpretation of rejected pre-consensus energy as anomaly evidence, with EDM serving as one density-oriented realization of Adapt.

\subsection{Training and Implementation Details}
\label{sec:training}
DDGAD has three stages: a graph convolutional autoencoder (AE) supplies initial embeddings $\vz^{(0)}$, an EDM denoiser implements node-wise Adapt, and ATC performs Combine and Fuse. The AE and EDM are trained offline without anomaly labels and remain fixed during inference; ATC introduces no additional trainable parameters.
The AE reconstructs node attributes and graph structure; because all nodes participate in this transductive objective, $\vz^{(0)}$ remains graph-conditioned and retains contextual influence introduced by graph encoding. Its loss and implementation details are reported in Appendix~\ref{app:experimental-details}.

\paragraph{Diffusion denoiser training.}
The Adapt operator is trained on frozen AE embeddings with the EDM objective~\citep{karras2022elucidating}
\begin{equation}
\mathcal{L}_{\mathrm{EDM}}=\E_{\vz^{(0)},\sigma,\epsilon}\!\left[\lambda(\sigma)
\left\|\mathcal{D}_\theta(\vz^{(0)}+\sigma\epsilon;\sigma)-\vz^{(0)}\right\|_2^2\right],
\label{eq:edm-loss}
\end{equation}
where $\epsilon\sim\mathcal N(0,\mI)$. Because the target is the observed latent population rather than a label-filtered normal subset, EDM provides a shared density-oriented adaptation rule whose node-wise application performs no new same-step neighbor aggregation. Complete architecture, preconditioning, and sampling details are in Appendix~\ref{app:edm}.

\subsection{Disagreement-based Anomaly Scoring}
\label{sec:scoring}
At each ATC step, DDGAD measures the pre-consensus disagreement between the node-wise Adapt estimate and the neighborhood-dependent contextual Combine estimate:
\begin{equation}
r_i^{(k)} \;=\; \|\psi_i^{(k)}-c_i^{(k)}\|_2 .
\label{eq:step-disagreement}
\end{equation}
Let $\mC^{(k)}$ and $\mR^{(k)}$ stack the Combine estimates and Adapt--Combine residuals, respectively. Then
\begin{equation}
\mC^{(k)}=\mW\boldsymbol{\Psi}^{(k)},\qquad
\mR^{(k)}=\boldsymbol{\Psi}^{(k)}-\mC^{(k)}
=(\mI-\mW)\boldsymbol{\Psi}^{(k)}.
\label{eq:residual-matrix}
\end{equation}
Here $\mW\boldsymbol{\Psi}^{(k)}$ is the local consensus formed by uniform neighborhood aggregation, so $(\mI-\mW)\boldsymbol{\Psi}^{(k)}$ is the part of the node-wise reference left unexplained by neighborhood consensus. The Fuse update further gives
\begin{equation}
\|\vz_i^{(k+1)}-\psi_i^{(k)}\|_2=(1-\alpha)r_i^{(k)}.
\label{eq:residual-correction}
\end{equation}
Thus, at iteration $k$, the distance moved from the node-wise Adapt reference toward neighborhood consensus is exactly proportional to $r_i^{(k)}$. A small residual indicates compatibility between the node-wise reference and its local neighborhood, whereas a large residual indicates a substantial conflict between them. Because the denoiser is stochastic and a single iteration may be transient, DDGAD averages disagreement over the full trajectory:
\begin{equation}
\mathrm{score}_i \;=\; \frac{1}{K}\sum_{k=0}^{K-1} r_i^{(k)} .
\label{eq:final-score}
\end{equation}
Persistent averaging retains repeated node--context incompatibility while reducing sensitivity to any one sampling step. The main detector uses only Eq.~\eqref{eq:final-score}; Table~\ref{tab:ablation} additionally examines its quadratic counterpart.

\Needspace{13\baselineskip}
\subsection{Algorithm}
\label{sec:algorithm}
Algorithm~\ref{ddgad-algorithm} summarizes ATC inference after the AE and EDM have been fixed.
\begin{algorithm}[H]
\caption{DDGAD inference}
\label{ddgad-algorithm}
\begin{algorithmic}[1]
\REQUIRE Graph $\gG=(\sV,\sE)$ with adjacency $\mA$, node features $\mX$, iterations $K$, $\alpha$
\ENSURE Anomaly scores for all $v_i\in\sV$
\STATE $\vz^{(0)}=\mathrm{GCN}(\mX,\mA)$; initialize $e_i=0$
\FOR{$k=0$ to $K-1$}
  \STATE \textbf{Adapt:} $\psi_i^{(k)}=\mathcal{D}_\theta(\vz_i^{(k)};\sigma_k,\boldsymbol{\xi}_i^{(k)})$
  \STATE \textbf{Combine:} compute $c_i^{(k)}$ by Eq.~\eqref{eq:combine}
  \STATE \textbf{Fuse:} $\vz_i^{(k+1)}=\alpha\psi_i^{(k)}+(1-\alpha)c_i^{(k)}$
  \STATE \textbf{Score:} $e_i\leftarrow e_i+\|\psi_i^{(k)}-c_i^{(k)}\|_2$
\ENDFOR
\STATE \textbf{Output:} $\mathrm{score}_i=e_i/K$
\end{algorithmic}
\end{algorithm}

\subsection{Theory: Rejected Consensus Energy and Structural Separation}
\label{sec:theory}
For an undirected graph, let $\mPi$ be the degree-stationary measure of $\mW$ and write its symmetric conjugate as $\mS=\mPi^{1/2}\mW\mPi^{-1/2}=\mU\boldsymbol{\Lambda}\mU^\top$. For eigenvector $\vu_\ell$ and eigenvalue $\lambda_\ell$, define the graph-frequency coefficient
\begin{equation}
\widehat{\boldsymbol{\Psi}}_\ell^{(k)}
=\vu_\ell^\top\mPi^{1/2}\boldsymbol{\Psi}^{(k)},
\label{eq:graph-frequency-main}
\end{equation}
and the weighted norm $\|\boldsymbol{Y}\|_{\mPi,F}^2=\operatorname{tr}(\boldsymbol{Y}^\top\mPi\boldsymbol{Y})$.

\begin{theorem}[Pre-consensus spectral energy]
\label{thm:spectral-energy}
The pre-consensus residual satisfies
\begin{equation}
\|\mR^{(k)}\|_{\mPi,F}^2
=\sum_{\ell=1}^{N}(1-\lambda_\ell)^2
\|\widehat{\boldsymbol{\Psi}}_\ell^{(k)}\|_2^2.
\label{eq:spectral-energy-main}
\end{equation}
\end{theorem}
Every component-wise consensus mode has $\lambda_\ell=1$ and contributes zero; modes rejected most strongly by neighborhood averaging receive the largest weight. At the graph level, Eq.~\eqref{eq:spectral-energy-main} identifies the $\mPi$-weighted aggregate residual energy rejected by consensus; it does not equate each node score $r_i^{(k)}$ with a localized spectral energy.

\begin{theorem}[Population separation under contextual mixing]
\label{thm:population-separation}
At step $k$, let $\mu_N^{(k)}$ and $\mu_A^{(k)}$ be normal and anomalous Adapt-space prototypes, with $D_k=\|\mu_A^{(k)}-\mu_N^{(k)}\|_2$. Suppose normal and detectable-anomaly neighborhoods contain at most anomaly proportions $\rho_N^{(k)}$ and $\rho_A^{(k)}$, respectively. If node-side estimates deviate from their class prototype by at most $\epsilon_{M,y}^{(k)}$ and contexts deviate from the corresponding two-population mixture by at most $\epsilon_{C,y}^{(k)}$, then any non-isolated detectable anomaly $a$ and non-isolated compatible normal node $n$ satisfy
\begin{equation}
r_a^{(k)}-r_n^{(k)}\geq
(1-\rho_A^{(k)}-\rho_N^{(k)})D_k
-\epsilon_{M,A}^{(k)}-\epsilon_{M,N}^{(k)}
-\epsilon_{C,A}^{(k)}-\epsilon_{C,N}^{(k)}.
\label{eq:population-separation-main}
\end{equation}
\end{theorem}
Thus $\rho_A+\rho_N<1$ defines a sufficient detectable regime rather than a universal graph property. Averaging Eq.~\eqref{eq:population-separation-main} over $K$ steps gives a positive score margin whenever its right-hand side has positive mean. Appendix~\ref{app:theory} proves the result and makes the context error degree-dependent under a sub-Gaussian sampling model.

The matched-source experiment also admits an exact population law. For a victim with neighborhood size $m>0$ and $\beta_r=r/(m+r)$, Appendix~\ref{app:theory} proves for squared disagreement that
\begin{equation}
\Delta_q(r)=A\beta_r+B\beta_r^2.
\label{eq:dose-law-main}
\end{equation}
Under the coefficient conditions stated in the appendix, anomalous sources yield a positive response that increases with dose while added degree is exactly matched. The appendix also gives a conditional input-to-state bound and concentration results for neighborhood estimation and stochastic averaging.

\section{Experiments}
\label{sec:experiments}
\subsection{Datasets}
We evaluate on six attributed-graph benchmarks: Books, Disney, Reddit, two Amazon co-purchasing variants, and Tolokers. Training is transductive and label-free, and scores are produced for all nodes before a fixed evaluation mask is applied. Dataset statistics, partitions, and implementation details are in Appendix~\ref{app:experimental-details}.

\subsection{Main results}
We compare with nineteen non-neural, reconstruction/generative, contrastive/self-supervised, and diffusion-based baselines on the same graph instances and evaluation partitions; implementation and score definitions are given in Appendix~\ref{app:experimental-details}.
\begin{table*}[t]
\centering
\caption{AUROC (\%) on six benchmarks. Every cell is reported as mean$\pm$sample standard deviation. The highest value is \textbf{bold} and the second-highest is underlined. ``Avg'' is the unweighted mean of the six dataset means.}
\label{tab:main}
\scriptsize
\setlength{\tabcolsep}{2pt}
\resizebox{\textwidth}{!}{%
\begin{tabular}{@{}lccccccc@{}}
\toprule
 & \textbf{Books} & \textbf{Disney} & \textbf{Reddit} & \textbf{Amazon-5KN} & \textbf{Amazon-Bal} & \textbf{Tolokers} & \textbf{Avg} \\
\midrule
\multicolumn{8}{l}{\emph{Traditional / shallow non-neural}} \\
LOF~\citep{breunig2000lof} & $37.96{\pm}8.22$ & $51.36{\pm}13.12$ & $56.84{\pm}2.29$ & $53.33{\pm}1.00$ & $54.52{\pm}1.32$ & $47.66{\pm}0.38$ & 50.28 \\
IF~\citep{liu2008isolation} & $57.58{\pm}5.26$ & $42.43{\pm}14.22$ & $55.57{\pm}1.32$ & $\mathbf{80.06{\pm}0.44}$ & $\mathbf{80.01{\pm}1.48}$ & $39.68{\pm}0.48$ & 59.22 \\
SCAN~\citep{xu2007scan} & $45.42{\pm}5.95$ & $51.65{\pm}10.86$ & $49.58{\pm}0.16$ & $50.00{\pm}0.00$ & $50.00{\pm}0.00$ & $53.81{\pm}0.31$ & 50.08 \\
Radar~\citep{li2017radar} & $58.17{\pm}11.69$ & $70.33{\pm}12.41$ & $44.82{\pm}1.87$ & $\underline{78.43{\pm}0.30}$ & $\underline{78.17{\pm}0.74}$ & $49.49{\pm}0.22$ & \underline{63.24} \\
\midrule
\multicolumn{8}{l}{\emph{GNN reconstruction / generative}} \\
GCN-AE~\citep{kipf2017semi} & $51.62{\pm}5.61$ & $39.13{\pm}19.97$ & $51.18{\pm}1.66$ & $26.40{\pm}1.02$ & $26.14{\pm}1.21$ & $45.84{\pm}1.24$ & 40.05 \\
DOMINANT~\citep{ding2019deep} & $38.51{\pm}6.76$ & $49.53{\pm}4.27$ & $55.65{\pm}0.99$ & $56.78{\pm}7.47$ & $56.10{\pm}6.82$ & $54.35{\pm}0.61$ & 51.82 \\
AnomalyDAE~\citep{fan2020anomalydae} & $42.95{\pm}7.25$ & $49.29{\pm}21.68$ & $48.97{\pm}0.54$ & $52.51{\pm}1.82$ & $45.98{\pm}10.08$ & $49.97{\pm}5.51$ & 48.28 \\
AdONE~\citep{bandyopadhyay2020outlier} & $43.25{\pm}1.56$ & $33.10{\pm}9.91$ & $47.85{\pm}2.17$ & $36.36{\pm}2.92$ & $35.99{\pm}2.59$ & $46.24{\pm}1.95$ & 40.47 \\
DONE~\citep{bandyopadhyay2020outlier} & $34.33{\pm}2.84$ & $38.89{\pm}11.36$ & $56.90{\pm}1.77$ & $35.94{\pm}0.27$ & $32.01{\pm}3.33$ & $48.96{\pm}1.45$ & 41.17 \\
GAD-NR~\citep{roy2024gadnr} & $38.78{\pm}5.96$ & $28.49{\pm}9.11$ & $55.51{\pm}1.38$ & $36.09{\pm}2.51$ & $35.19{\pm}1.76$ & $\underline{55.74{\pm}0.89}$ & 41.63 \\
GAAN~\citep{chen2020generative} & $46.99{\pm}6.84$ & $48.94{\pm}22.06$ & $49.51{\pm}2.80$ & $53.49{\pm}0.85$ & $53.25{\pm}0.67$ & $51.72{\pm}0.63$ & 50.65 \\
\midrule
\multicolumn{8}{l}{\emph{Contrastive / self-supervised}} \\
CoLA~\citep{liu2021anomaly} & $43.17{\pm}12.04$ & $50.24{\pm}7.44$ & $51.33{\pm}1.30$ & $38.17{\pm}0.47$ & $37.80{\pm}1.30$ & $51.57{\pm}0.34$ & 45.38 \\
GraphMAE~\citep{hou2022graphmae} & $61.85{\pm}4.45$ & $\underline{76.60{\pm}12.77}$ & $48.70{\pm}5.15$ & $73.70{\pm}0.64$ & $74.59{\pm}0.61$ & $43.08{\pm}0.40$ & 63.09 \\
CONAD~\citep{xu2022conad} & $38.39{\pm}6.81$ & $47.52{\pm}7.58$ & $56.11{\pm}0.96$ & $60.09{\pm}11.02$ & $56.05{\pm}6.80$ & $54.55{\pm}0.80$ & 52.12 \\
OCGNN~\citep{wang2021ocgnn} & $53.10{\pm}8.10$ & $74.23{\pm}1.68$ & $49.66{\pm}7.66$ & $51.94{\pm}1.13$ & $52.97{\pm}2.22$ & $44.19{\pm}0.65$ & 54.35 \\
PREM~\citep{pan2023prem} & $33.68{\pm}7.02$ & $24.00{\pm}10.46$ & $57.02{\pm}1.42$ & $41.56{\pm}7.59$ & $41.16{\pm}6.19$ & $47.58{\pm}2.05$ & 40.83 \\
GADAM~\citep{chen2024gadam} & $49.91{\pm}7.23$ & $28.61{\pm}19.12$ & $\underline{58.03{\pm}1.76}$ & $30.42{\pm}4.03$ & $31.21{\pm}4.49$ & $42.89{\pm}1.19$ & 40.18 \\
TAM~\citep{qiao2023truncated} & $45.11{\pm}8.26$ & $40.54{\pm}11.09$ & $\mathbf{58.20{\pm}2.56}$ & $61.33{\pm}1.20$ & $62.91{\pm}0.31$ & $40.30{\pm}0.57$ & 51.40 \\
\midrule
\multicolumn{8}{l}{\emph{Diffusion-based}} \\
DiffGAD~\citep{li2025diffgad} & $\mathbf{69.10{\pm}6.42}$ & $52.96{\pm}20.06$ & $56.49{\pm}0.63$ & $49.69{\pm}2.36$ & $48.11{\pm}3.06$ & $54.39{\pm}2.17$ & 55.12 \\
\midrule
\multicolumn{8}{l}{\emph{Ours}} \\
\textbf{DDGAD (ours)} & $\underline{62.18{\pm}8.11}$ & $\mathbf{80.85{\pm}9.76}$ & $57.29{\pm}2.14$ & $69.98{\pm}3.60$ & $67.02{\pm}0.74$ & $\mathbf{62.53{\pm}0.24}$ & \textbf{66.64} \\
\bottomrule
\end{tabular}
}
\end{table*}

Over five seeds, DDGAD achieves the highest average AUROC of $66.64\%$ across six benchmarks, exceeding the next-highest average, obtained by Radar, by $3.40$ percentage points. It ranks first on Disney and Tolokers and second on Books. Architecture and reporting details are in Appendix~\ref{app:experimental-details}.

\subsection{Ablation Study}
Table~\ref{tab:ablation} isolates label-free inference signals using one shared AE/EDM checkpoint per dataset.

\begin{table}[t]
\centering
\caption{Signal ablation (AUROC, \%). Here $r_i$ is Eq.~\eqref{eq:final-score} and $q_i=K^{-1}\sum_{k=0}^{K-1}\|\psi_i^{(k)}-c_i^{(k)}\|_2^2$.}
\label{tab:ablation}
\scriptsize
\setlength{\tabcolsep}{3pt}
\begin{tabular}{llcc}
\toprule
\textbf{Variant} & \textbf{Score used} & \textbf{Books} & \textbf{Disney} \\
\midrule
AE only                  & reconstruction error & 51.25 & 39.97 \\
AE + EDM                 & denoising deviation  & 55.88 & 35.31 \\
Persistent disagreement & $r_i$                & 64.04 & 71.75 \\
Quadratic disagreement  & $q_i$                & 64.17 & 72.18 \\
\bottomrule
\end{tabular}
\end{table}

EDM deviation does not consistently improve AE scoring, whereas both disagreement summaries are substantially stronger. The quadratic variant is only marginally higher than persistent disagreement, by $0.13$ AUROC on Books and $0.43$ on Disney. We retain the linear persistent score in Eq.~\eqref{eq:final-score} as the default because it preserves the scale of each per-iteration Adapt--Combine discrepancy and is less sensitive to isolated large deviations. Their comparable performance identifies Adapt--Combine disagreement, rather than a particular trajectory aggregation form, as the core signal.

\subsection{Hyperparameter Sensitivity}
\begin{figure}[!htb]
  \centering
  \includegraphics[width=0.92\linewidth]{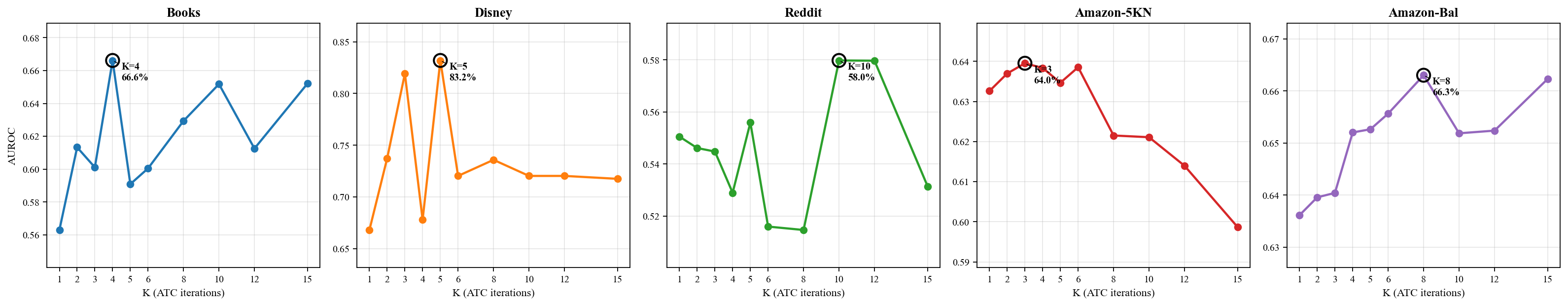}
  \caption{AUROC versus ATC iteration count $K$ on each dataset.}
  \label{fig:hparam-K}
\end{figure}

Figure~\ref{fig:hparam-K} traces AUROC across ATC iteration counts $K$ for each dataset. The dataset-dependent response shows that persistence is an active component of the detector rather than an inert repetition count.

\subsection{Controlled Context-Source Intervention}
\label{sec:context-intervention}
We test whether disagreement responds to the \emph{source} of contextual evidence rather than merely to adding edges. For each Amazon graph, we compute the clean encoder output $\vz^{(0)}$ once and freeze both it and all EDM reference snapshots from the clean trajectory. We then change only the adjacency consumed by a uniform Combine operator. In each of 30 randomized trials, 500 normal victims receive $r\in\{1,2,4,8,16\}$ distinct incoming sources sampled either from anomalous nodes or from normal nodes. The paired endpoint is
\begin{equation}
\Delta_{\mathrm{src}}(r)
=s_{\mathrm{anomaly\text{-}add}}(r)-s_{\mathrm{normal\text{-}add}}(r),
\label{eq:source-intervention}
\end{equation}
so both counterfactuals add the same number of edges. Labels are used only to construct this post-hoc intervention, never for training or score construction. We restrict victims to nodes with non-empty pre-intervention neighborhoods: adding the first neighbor to an isolated node changes the existence of the contextual reference, rather than only its source composition.

\begin{figure}[H]
  \centering
  \includegraphics[width=0.92\textwidth]{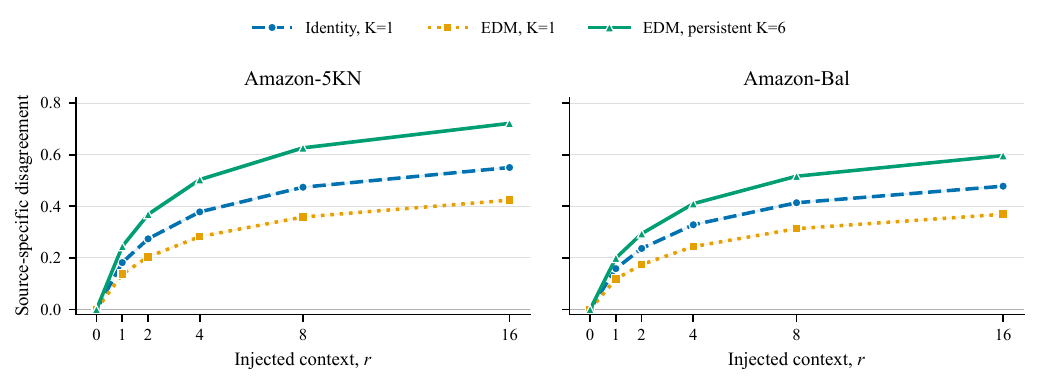}
  \caption{Matched post-encoder source intervention validating source-specific disagreement response. Curves show mean $\Delta_{\mathrm{src}}(r)$ over 30 randomized trials with 500 paired victims per trial; shaded regions are 95\% trial-level paired-bootstrap confidence intervals. The clean encoder output and EDM reference trajectory are frozen, and only uniform Combine receives the intervention edges.}
  \label{fig:context-dose-response}
\end{figure}

Figure~\ref{fig:context-dose-response} reports the complete dose response. Identity already yields a positive source effect, showing that the principle extends beyond diffusion, while persistent EDM is strongest throughout the tested range. The response increases monotonically on both graphs, and all 30 trials have positive dose slopes for every displayed method. Table~\ref{tab:context-intervention} reports the $r=16$ matched-source effects and the uncertainty and consistency statistics for persistent EDM.

\begin{table}[H]
\centering
\caption{Matched context-source intervention at $r=16$. The three method columns report $\Delta_{\mathrm{src}}(16)$ from Eq.~\eqref{eq:source-intervention}, not accuracy. Brackets give the persistent-EDM 95\% trial-level paired-bootstrap interval from 50,000 resamples. The remaining columns report the standardized persistent-EDM effect, positive-victim fraction, and exact two-sided Wilcoxon signed-rank $p$-value over 30 trials; both datasets have 30/30 positive trial-level effects.}
\label{tab:context-intervention}
\scriptsize
\setlength{\tabcolsep}{4pt}
\begin{tabular}{lcccccc}
\toprule
& \multicolumn{3}{c}{\textbf{Matched-source effect $\Delta_{\mathrm{src}}(16)$}} & \multicolumn{3}{c}{\textbf{Persistent-EDM statistics}} \\
\cmidrule(lr){2-4}\cmidrule(lr){5-7}
\textbf{Dataset} & \textbf{Identity $K=1$} & \textbf{EDM $K=1$} & \textbf{EDM $K=6$ [95\% CI]} & \textbf{$\Delta_{\mathrm{src}}/\sigma_{\mathrm{clean,N}}$} & \textbf{Victim $+$} & \textbf{$p$} \\
\midrule
Amazon-5KN & 0.5505 & 0.4240 & \textbf{0.7217 [0.7155, 0.7280]} & 0.448 & 98.0\% & $1.86\!\times\!10^{-9}$ \\
Amazon-Bal & 0.4781 & 0.3693 & \textbf{0.5966 [0.5921, 0.6012]} & 0.343 & 98.3\% & $1.86\!\times\!10^{-9}$ \\
\bottomrule
\end{tabular}
\end{table}

With node-side information fixed and added degree matched, anomalous sources induce consistently larger Adapt--Combine disagreement than normal sources. The stronger persistent-EDM effect further supports the source-sensitivity of disagreement beyond additional connectivity.

\paragraph{Is diffusion necessary for Adapt?}
To test whether effective node--context disagreement relies on diffusion, we replace EDM with Identity, Linear, and MLP Adapt operators while keeping the frozen AE initialization, uniform Combine, disagreement score, and ATC iterations $K=5$ fixed.

\begin{table}[H]
\centering
\caption{Detection performance--efficiency trade-off of $K=5$ Adapt controls on Disney. AUROC/AP are percentages. Time is median complete-ATC latency on an RTX 4070; parameter counts exclude the shared frozen AE. All rows share one frozen AE checkpoint and uniform Combine.}
\label{tab:adapt-ablation}
\scriptsize
\begin{tabular}{lrrrr}
\toprule
\textbf{Adapt Operator} & \textbf{AUROC} & \textbf{AP} & \textbf{Time (ms)} & \textbf{Adapt Params} \\
\midrule
Identity & 74.82 & 7.74 & 1.31 & 0 \\
Linear & 79.43 & 9.15 & 1.23 & 272 \\
MLP & 79.08 & 8.93 & 1.59 & 2,128 \\
\textbf{EDM}~\citep{karras2022elucidating} & \textbf{85.11} & \textbf{14.39} & 134.65 & 471,056 \\
\bottomrule
\end{tabular}
\end{table}

All non-diffusion Adapt controls produce effective detection signals, indicating that Adapt--Combine disagreement is not unique to EDM. Identity already achieves 74.82\% AUROC, while Linear and MLP both approach 79\% AUROC. EDM performs best with 85.11\% AUROC and 14.39\% AP, improving over the strongest non-diffusion control at substantially higher computational cost. Thus, diffusion is not necessary for informative disagreement, but provides a stronger, albeit more expensive, Adapt realization.

\subsection{Efficiency and Limitations}
With latent width $d_z$, denoiser width $h_D$, depth $L_D$, and $S_D$ solver steps, one EDM-based ATC iteration costs $O(S_DNL_Dh_D^2+|\mathcal{E}|d_z)$ up to input/output projections; uniform Combine is the edge-dependent term. On Disney, the complete $K=5$ loop takes $134.65$ ms with EDM and $1.59$ ms with MLP.

DDGAD exposes rather than removes upstream contamination: coherent anomalous neighborhoods can yield little disagreement as $\rho_A\!\to\!1$, and degree-zero nodes have zero contextual residual.

\section{Conclusion}
DDGAD reinterprets pre-consensus node--context disagreement as anomaly evidence. Theory characterizes its spectral, structural, and source-response properties, while benchmarks, matched-source interventions, and Adapt controls support the mechanism. The signal is not diffusion-specific, although EDM provides the strongest tested realization. Overall, rejected consensus error offers an informative and interpretable basis for GAD.

\paragraph{AI use statement.}
Generative AI tools assisted with \LaTeX{} migration, language editing, and
compilation troubleshooting. The authors reviewed the resulting manuscript and
take responsibility for its content, claims, and artifacts.

\bibliography{references}
\bibliographystyle{iclr2027_conference}

\clearpage
\appendix
\section{Theoretical Analysis of ATC Dynamics}
\label{app:theory}

We analyze the object implemented by DDGAD: a stochastic node-wise diffusion Adapt operator, a uniform neighborhood Combine operator, and their persistent pre-consensus residual. The results establish structural isolation, a mixture-aware manifold interpretation, an exact spectral characterization of rejected consensus energy, conditional dynamics under forcing, population separation under contextual mixing, a matched-source dose law, and concentration of neighborhood estimation and persistent averaging.

\subsection{Propagation Before ATC and Isolation Inside Adapt}

\begin{prop}[Propagation support in a linearized GCN]
Let an $L$-layer linearized GCN be $\mH^{(L)}=\widetilde{\mA}^{L}\mX\mW_{1:L}$, where $\widetilde{\mA}$ is the normalized adjacency with self-loops and $\mW_{1:L}$ is the composed feature map. For a perturbation $\Delta\mX$ supported on node set $\sA$,
\begin{equation}
\Delta\mH^{(L)}=\widetilde{\mA}^{L}\Delta\mX\mW_{1:L},\qquad
\Delta\vh_i^{(L)}=\sum_{j\in\sA}(\widetilde{\mA}^{L})_{ij}\Delta\vx_j\mW_{1:L}.
\label{eq:app-gcn-propagation}
\end{equation}
Therefore node $i$ is unaffected whenever no walk of length at most $L$ in the original graph connects it to the perturbed support; equivalently, no length-$L$ walk connects them in the self-loop-augmented graph.
\end{prop}

\begin{Proof}
Substituting $\mX+\Delta\mX$ into the linearized encoder and subtracting the unperturbed output gives the first equality. Expanding its $i$th row and using the support of $\Delta\mX$ gives the second. The result characterizes where contamination can propagate; it does not imply monotonic growth with depth.
\end{Proof}

\begin{prop}[Conditional cross-node isolation of Adapt]
Condition on the EDM sampling noises $\{\boldsymbol{\xi}_i^{(k)}\}_{i=1}^{N}$. Because the shared Adapt map is applied node by node,
\begin{equation}
\frac{\partial\psi_i^{(k)}}{\partial\vz_j^{(k)}}
=\mathbf{0}_{d_z\times d_z},
\qquad j\neq i.
\label{eq:app-adapt-jacobian}
\end{equation}
If two latent states differ only on a node set $S$, their adapted outputs, under coupled equal noises, differ only on $S$.
\end{prop}

\begin{Proof}
The input to $\psi_i^{(k)}=\mathcal D_\theta(\vz_i^{(k)};\sigma_k,\boldsymbol{\xi}_i^{(k)})$ contains no $\vz_j^{(k)}$ for $j\neq i$; hence Eq.~\eqref{eq:app-adapt-jacobian} holds. Equal node-wise inputs and coupled noises outside $S$ then produce equal outputs. This is a same-step conditional-isolation statement: Adapt performs no new cross-node aggregation, but for $k\geq1$ its input $\vz_i^{(k)}$ may already contain neighbor information introduced by the preceding Fuse.
\end{Proof}

\subsection{Mixture-Aware Manifold Interpretation of Diffusion Adapt}

The denoising--score identity used below follows the standard connection between denoising objectives and the score of a smoothed data density~\citep{vincent2011connection,alain2014regularized}. Let the frozen AE latent distribution be the mixture
\begin{equation}
p(\vz)=(1-\pi_A)p_N(\vz)+\pi_A p_A(\vz),
\label{eq:app-latent-mixture}
\end{equation}
where the components denote normal and anomalous latent populations only for analysis; these labels are not used by training. Let $p_{N,\sigma}$, $p_{A,\sigma}$, and $p_\sigma$ be their Gaussian-smoothed densities.

\begin{prop}[Diffusion score and latent-mixture bias]
For $\vy=\vz+\sigma\boldsymbol{\epsilon}$ with $\boldsymbol{\epsilon}\sim\mathcal N(0,\mI)$, the population-optimal MSE denoiser is
\begin{equation}
\mathcal D_\sigma^*(\vy)
=\E[\vz\mid\vy]
=\vy+\sigma^2\nabla_{\vy}\log p_\sigma(\vy).
\label{eq:app-tweedie}
\end{equation}
Define the local posterior anomaly mass and component scores by
\begin{equation}
\omega_A(\vy)=\frac{\pi_A p_{A,\sigma}(\vy)}{p_\sigma(\vy)},\qquad
s_q(\vy)=\nabla_{\vy}\log p_{q,\sigma}(\vy),\quad q\in\{N,A\}.
\label{eq:app-posterior-mass}
\end{equation}
Then
\begin{equation}
\nabla\log p_\sigma(\vy)
=(1-\omega_A(\vy))s_N(\vy)+\omega_A(\vy)s_A(\vy).
\label{eq:app-mixture-score}
\end{equation}
If the learned denoiser has pointwise error at most $\epsilon_\theta(\vy,\sigma)$ from $\mathcal D_\sigma^*$, its deviation from the normal-component denoiser obeys
\begin{equation}
\|\mathcal D_\theta(\vy;\sigma)-\mathcal D_{N,\sigma}^*(\vy)\|_2
\leq\epsilon_\theta(\vy,\sigma)+\sigma^2\omega_A(\vy)\|s_A(\vy)-s_N(\vy)\|_2.
\label{eq:app-mixture-bias}
\end{equation}
\end{prop}

\begin{Proof}
Tweedie's identity for additive Gaussian corruption gives Eq.~\eqref{eq:app-tweedie}. Differentiating the log-density of the smoothed mixture yields Eq.~\eqref{eq:app-mixture-score}. Finally, subtracting $\mathcal D_{N,\sigma}^*(\vy)=\vy+\sigma^2s_N(\vy)$ from the mixture denoiser and applying the triangle inequality with the learned approximation error yields Eq.~\eqref{eq:app-mixture-bias}.
\end{Proof}

\begin{corollary}[Conditional approach to a dominant manifold]
Let $\sM_N$ be a dominant normal latent manifold and suppose the ideal normal-component denoiser satisfies
\begin{equation}
\operatorname{dist}(\mathcal D_{N,\sigma}^*(\vy),\sM_N)
\leq q_\sigma\operatorname{dist}(\vy,\sM_N)+\epsilon_M(\sigma),\qquad 0\leq q_\sigma<1.
\label{eq:app-normal-manifold-contraction}
\end{equation}
Then
\begin{equation}
\operatorname{dist}(\mathcal D_\theta(\vy;\sigma),\sM_N)
\leq q_\sigma\operatorname{dist}(\vy,\sM_N)+\epsilon_M(\sigma)+\epsilon_\theta(\vy,\sigma)
+\sigma^2\omega_A(\vy)\|s_A(\vy)-s_N(\vy)\|_2.
\label{eq:app-manifold-bound}
\end{equation}
\end{corollary}

\begin{Proof}
Distance to a closed set is $1$-Lipschitz. Applying this fact to Eq.~\eqref{eq:app-mixture-bias} and then substituting Eq.~\eqref{eq:app-normal-manifold-contraction} proves the claim.
\end{Proof}

The corollary formalizes the limited sense in which Adapt is manifold-oriented. A high global anomaly ratio is not by itself decisive; the bias is governed by the local posterior mass $\omega_A(\vy)$. Conversely, a coherent anomalous mode can make this term large, so diffusion is not assumed to be a clean-manifold projector. Multi-step EDM sampling composes such denoising fields; its network and numerical errors are collected in the Adapt approximation term used below.

\subsection{Inverted Consensus and Pre-consensus Spectral Energy}

Let $\mW$ be the fixed row-normalized neighborhood matrix,
\begin{equation}
W_{ij}=\begin{cases}
1/|\sN(i)|, & j\in\sN(i),\ |\sN(i)|>0,\\
1, & j=i,\ |\sN(i)|=0,\\
0, & \text{otherwise}.
\end{cases}
\label{eq:app-uniform-weight}
\end{equation}
Let $\boldsymbol{\Psi}^{(k)}$ stack the adapted states. Then
\begin{equation}
\mC^{(k)}=\mW\boldsymbol{\Psi}^{(k)},
\qquad
\mR^{(k)}=(\mI-\mW)\boldsymbol{\Psi}^{(k)}.
\label{eq:app-matrix-residual}
\end{equation}

\begin{prop}[Local estimation and correction identity]
For a non-isolated node, the DDGAD update is the unique minimizer
\begin{equation}
\vz_i^{(k+1)}=\arg\min_{\vu\in\R^{d_z}}\Bigl[\alpha\|\vu-\psi_i^{(k)}\|_2^2
+\frac{1-\alpha}{|\sN(i)|}\sum_{j\in\sN(i)}\|\vu-\psi_j^{(k)}\|_2^2\Bigr].
\label{eq:app-local-objective}
\end{equation}
Furthermore,
\begin{equation}
\|\vz_i^{(k+1)}-\psi_i^{(k)}\|_2
=(1-\alpha)\|c_i^{(k)}-\psi_i^{(k)}\|_2
=(1-\alpha)r_i^{(k)}.
\label{eq:app-correction-identity}
\end{equation}
\end{prop}

\begin{Proof}
The first-order condition for the strictly convex objective in Eq.~\eqref{eq:app-local-objective} gives the unique minimizer $\alpha\psi_i^{(k)}+(1-\alpha)c_i^{(k)}$. Subtracting $\psi_i^{(k)}$ from this minimizer yields Eq.~\eqref{eq:app-correction-identity}. For an isolated node, Eq.~\eqref{eq:app-uniform-weight} gives $c_i^{(k)}=\psi_i^{(k)}$, so the residual is zero.
\end{Proof}

Because $c_i^{(k)}$ is a convex combination of adapted neighbors and $\vz_i^{(k+1)}$ is a convex combination of $\psi_i^{(k)}$ and $c_i^{(k)}$, bounded Adapt outputs imply bounded ATC states. We therefore use this exact boundedness statement instead of claiming geometric convergence of the stochastic, stateful map.

For the undirected graph, set $m_i^+=\max\{|\sN(i)|,1\}$, $\varpi_i=m_i^+/\sum_jm_j^+$, and $\mPi=\operatorname{diag}(\boldsymbol{\varpi})$. The isolated-node convention in Eq.~\eqref{eq:app-uniform-weight} makes $\mW$ reversible:
\begin{equation}
\mPi\mW=\mW^\top\mPi.
\label{eq:app-detailed-balance}
\end{equation}
Hence $\mS=\mPi^{1/2}\mW\mPi^{-1/2}$ is symmetric. Write $\mS=\mU\boldsymbol{\Lambda}\mU^\top$, with eigenvalues $\lambda_\ell\in[-1,1]$, and define the row-vector graph-frequency coefficient
\begin{equation}
\widehat{\boldsymbol{\Psi}}_\ell^{(k)}
=\vu_\ell^\top\mPi^{1/2}\boldsymbol{\Psi}^{(k)}.
\label{eq:app-graph-frequency}
\end{equation}

\begin{theorem}[Pre-consensus spectral characterization]
For $\|\boldsymbol{Y}\|_{\mPi,F}^2=\operatorname{tr}(\boldsymbol{Y}^\top\mPi\boldsymbol{Y})$,
\begin{equation}
\|\mR^{(k)}\|_{\mPi,F}^2
=\sum_{\ell=1}^{N}(1-\lambda_\ell)^2
\|\widehat{\boldsymbol{\Psi}}_\ell^{(k)}\|_2^2.
\label{eq:app-spectral-energy}
\end{equation}
\end{theorem}

\begin{Proof}
Using reversibility and Eq.~\eqref{eq:app-matrix-residual},
\begin{equation}
\mPi^{1/2}\mR^{(k)}
=(\mI-\mS)\mPi^{1/2}\boldsymbol{\Psi}^{(k)}
=\mU(\mI-\boldsymbol{\Lambda})\mU^\top
\mPi^{1/2}\boldsymbol{\Psi}^{(k)}.
\end{equation}
Taking squared Frobenius norms and using the orthogonality of $\mU$ yields Eq.~\eqref{eq:app-spectral-energy}.
\end{Proof}

Every eigenvector with $\lambda_\ell=1$ is a component-wise consensus mode and contributes zero. Components most strongly rejected by the averaging operator receive weight $(1-\lambda_\ell)^2$. Thus, in the $\mPi$-weighted graph-level aggregate, the residual energy is exactly the pre-consensus spectral energy in Eq.~\eqref{eq:app-spectral-energy}; the theorem does not assign this identity separately to each node score.

\subsection{Persistent ATC Dynamics under Forcing}

Consider one connected component, renormalize $\boldsymbol{\varpi}$ on that component, and let $\mJ=\mathbf{1}\boldsymbol{\varpi}^\top$ be its consensus projector. Define
\begin{equation}
\mB=\alpha\mI+(1-\alpha)\mW,
\qquad
\rho_\perp(\mB)=\max_{\ell\geq2}|\alpha+(1-\alpha)\lambda_\ell|.
\label{eq:app-consensus-operator}
\end{equation}
Write the stochastic Adapt output as
\begin{equation}
\boldsymbol{\Psi}^{(k)}=\mathcal F_k(\mZ^{(k)})+\boldsymbol{\Upsilon}^{(k)},
\label{eq:app-forcing-decomposition}
\end{equation}
where the shared row-wise map $\mathcal F_k$ is $L_k$-Lipschitz in $\|\cdot\|_{\mPi,F}$ and $\boldsymbol{\Upsilon}^{(k)}$ collects sampling noise, approximation error, and node- or source-specific forcing.

\begin{theorem}[Conditional input-to-state bound]
Let $\boldsymbol{\delta}^{(k)}=(\mI-\mJ)\mZ^{(k)}$. The Fuse dynamics satisfy
\begin{equation}
\|\boldsymbol{\delta}^{(k+1)}\|_{\mPi,F}
\leq q_k\|\boldsymbol{\delta}^{(k)}\|_{\mPi,F}+u_k,
\quad
q_k=L_k\rho_\perp(\mB),
\quad
u_k=\rho_\perp(\mB)\|(\mI-\mJ)\boldsymbol{\Upsilon}^{(k)}\|_{\mPi,F}.
\label{eq:app-iss-one-step}
\end{equation}
If $q_k\leq q<1$ and $u_k\leq\bar u$, then
\begin{equation}
\|\boldsymbol{\delta}^{(k)}\|_{\mPi,F}
\leq q^k\|\boldsymbol{\delta}^{(0)}\|_{\mPi,F}
+\left(\frac{1-q^k}{1-q}\right)\bar u.
\label{eq:app-iss-unrolled}
\end{equation}
\end{theorem}

\begin{Proof}
Because $\mB$ commutes with $\mJ$ and has weighted operator norm $\rho_\perp(\mB)$ on the disagreement subspace,
\begin{equation}
\|(\mI-\mJ)\mB\mX\|_{\mPi,F}
\leq\rho_\perp(\mB)\|(\mI-\mJ)\mX\|_{\mPi,F}.
\end{equation}
Moreover, $\mathcal F_k(\mJ\mZ)$ is a consensus matrix. Projection optimality and Lipschitzness therefore give $\|(\mI-\mJ)\mathcal F_k(\mZ)\|_{\mPi,F}\leq L_k\|(\mI-\mJ)\mZ\|_{\mPi,F}$. Substitution into $\mZ^{(k+1)}=\mB\boldsymbol{\Psi}^{(k)}$ yields Eq.~\eqref{eq:app-iss-one-step}, and induction yields Eq.~\eqref{eq:app-iss-unrolled}.
\end{Proof}

This conditional stability result characterizes contraction under compatible dynamics; stochastic or source-specific inputs determine the residual floor, and persistent anomaly evidence requires non-canceling forcing across iterations.

\subsection{Population Separation under Contextual Mixing}

At step $k$, let $\mu_N^{(k)},\mu_A^{(k)}\in\mathbb R^{d_z}$ be normal and anomalous Adapt-space prototypes and define
\begin{equation}
D_k=\|\mu_A^{(k)}-\mu_N^{(k)}\|_2.
\label{eq:app-prototype-distance}
\end{equation}
For node $i$, let $\rho_i^{(k)}$ denote the anomalous population mass in its neighborhood and let
\begin{equation}
\bar c_i^{(k)}=(1-\rho_i^{(k)})\mu_N^{(k)}+\rho_i^{(k)}\mu_A^{(k)}
\label{eq:app-mixture-context}
\end{equation}
be the corresponding two-population context. This is a reference model, not an assumption that every embedding equals its class center: node heterogeneity, finite-neighborhood error, and mismatch from two-population mixing appear explicitly below.

\begin{theorem}[Perturbed population separation]
\label{thm:app-population-separation}
Let $a$ be a non-isolated anomaly in the detectable regime and $n$ a non-isolated compatible normal node. Suppose
\begin{align}
\rho_a^{(k)}&\leq\rho_A^{(k)}, &
\rho_n^{(k)}&\leq\rho_N^{(k)},\label{eq:app-rho-bounds}\\
\|\psi_i^{(k)}-\mu_{y_i}^{(k)}\|_2&\leq\epsilon_{M,y_i}^{(k)}, &
\|c_i^{(k)}-\bar c_i^{(k)}\|_2&\leq\epsilon_{C,y_i}^{(k)}
\label{eq:app-prototype-errors}
\end{align}
for $i\in\{a,n\}$ and $y_i\in\{A,N\}$. Then
\begin{equation}
r_a^{(k)}-r_n^{(k)}\geq\Gamma_k,
\quad
\Gamma_k=(1-\rho_A^{(k)}-\rho_N^{(k)})D_k
-\epsilon_{M,A}^{(k)}-\epsilon_{M,N}^{(k)}
-\epsilon_{C,A}^{(k)}-\epsilon_{C,N}^{(k)}.
\label{eq:app-population-margin}
\end{equation}
Consequently,
\begin{equation}
\mathrm{score}_a-\mathrm{score}_n
\geq\frac{1}{K}\sum_{k=0}^{K-1}\Gamma_k,
\label{eq:app-persistent-population-margin}
\end{equation}
and the two nodes are strictly ordered whenever the right-hand side is positive.
\end{theorem}

\begin{Proof}
For the anomaly, the reverse triangle inequality and Eqs.~\eqref{eq:app-mixture-context}--\eqref{eq:app-prototype-errors} give
\begin{equation}
r_a^{(k)}\geq(1-\rho_a^{(k)})D_k
-\epsilon_{M,A}^{(k)}-\epsilon_{C,A}^{(k)}
\geq(1-\rho_A^{(k)})D_k
-\epsilon_{M,A}^{(k)}-\epsilon_{C,A}^{(k)}.
\end{equation}
For the normal node, the triangle inequality gives
\begin{equation}
r_n^{(k)}\leq\rho_n^{(k)}D_k
+\epsilon_{M,N}^{(k)}+\epsilon_{C,N}^{(k)}
\leq\rho_N^{(k)}D_k
+\epsilon_{M,N}^{(k)}+\epsilon_{C,N}^{(k)}.
\end{equation}
Subtracting the two bounds yields Eq.~\eqref{eq:app-population-margin}. Averaging over $k$ then yields Eq.~\eqref{eq:app-persistent-population-margin}.
\end{Proof}

The condition $\rho_A^{(k)}+\rho_N^{(k)}<1$ is sufficient, not necessary, and defines a structurally detectable regime for non-isolated nodes. Isolated nodes are excluded because Eq.~\eqref{eq:app-uniform-weight} sets $c_i^{(k)}=\psi_i^{(k)}$ and hence $r_i^{(k)}=0$. Moreover, as a coherent anomalous community drives $\rho_A^{(k)}\to1$, its guaranteed margin vanishes. These are expected failure modes of a detector based only on node--context disagreement.

\begin{corollary}[Degree-dependent Combine error]
\label{cor:app-degree-context}
Assume all nodes are non-isolated and write $m_i=|\sN(i)|$. Conditional on neighbor types, suppose the centered neighbor Adapt outputs are independent and each coordinate of $\psi_j^{(k)}-\mu_{y_j}^{(k)}$ is $\sigma_C$-sub-Gaussian. Then for any $\delta_C\in(0,1)$, with probability at least $1-\delta_C$, simultaneously for all $N$ nodes and $K$ steps,
\begin{equation}
\|c_i^{(k)}-\bar c_i^{(k)}\|_2
\leq\sigma_C\sqrt{\frac{2d_z\log(2NKd_z/\delta_C)}{m_i}}.
\label{eq:app-degree-context-bound}
\end{equation}
Any additional deviation from the two-population reference adds as a structural-mismatch term to the right-hand side.
\end{corollary}

\begin{Proof}
For a fixed node, step, and coordinate, the average of $m_i$ conditionally independent $\sigma_C$-sub-Gaussian variables has tail probability at most $2\exp(-m_i t^2/(2\sigma_C^2))$. Applying a union bound over the $NKd_z$ events and using $\|x\|_2\leq\sqrt{d_z}\|x\|_\infty$ yields Eq.~\eqref{eq:app-degree-context-bound}.
\end{Proof}

Corollary~\ref{cor:app-degree-context} makes the role of degree explicit while retaining the unavoidable $d_z$ dependence of a high-dimensional Euclidean-norm bound. It is an idealized finite-sample statement: correlated neighbors or heterogeneous class-conditional means require an effective sample size or an added mismatch term.

\subsection{Matched-Source Dose Response}

Consider a non-isolated victim with original neighborhood size $m$, base context $c_0$, and fixed node-side reference $\psi$. Add $r$ sources with normal population mean $\mu_N$ or anomalous mean $\mu_A=\mu_N+\Delta\mu$. With $\beta_r=r/(m+r)$,
\begin{equation}
c_N(r)=(1-\beta_r)c_0+\beta_r\mu_N,
\qquad
c_A(r)=c_N(r)+\beta_r\Delta\mu.
\label{eq:app-matched-contexts}
\end{equation}

\begin{theorem}[Degree-controlled source-response law]
For $q_N(r)=\|\psi-c_N(r)\|_2^2$ and $q_A(r)=\|\psi-c_A(r)\|_2^2$,
\begin{equation}
\Delta_q(r):=q_A(r)-q_N(r)
=A\beta_r+B\beta_r^2,
\label{eq:app-dose-law}
\end{equation}
where
\begin{equation}
A=-2\langle\psi-c_0,\Delta\mu\rangle,\qquad
B=\|\Delta\mu\|_2^2+2\langle\mu_N-c_0,\Delta\mu\rangle.
\label{eq:app-dose-coefficients}
\end{equation}
If $A\geq0$, $B\geq0$, and $(A,B)\neq(0,0)$, then $\Delta_q(r)>0$ and is strictly increasing for $r>0$.
\end{theorem}

\begin{Proof}
Substituting the matched contexts from Eq.~\eqref{eq:app-matched-contexts} gives
\begin{align}
\Delta_q(r)
&=\beta_r^2\|\Delta\mu\|_2^2
-2\beta_r\langle\psi-c_N(r),\Delta\mu\rangle\\
&=-2\beta_r\langle\psi-c_0,\Delta\mu\rangle
+\beta_r^2\!\left(\|\Delta\mu\|_2^2+2\langle\mu_N-c_0,\Delta\mu\rangle\right).
\end{align}
The resulting expression is Eq.~\eqref{eq:app-dose-law}. Under the stated conditions, $\beta_r$ is strictly increasing in $r$ and $A+2B\beta_r>0$; hence the response is positive and strictly increasing.
\end{Proof}

The theorem explains why equal added degree is essential: both counterfactuals share the same $\beta_r$, leaving source composition as the difference. For a persistent quadratic score, summing Eq.~\eqref{eq:app-dose-law} over steps preserves positivity when the conditions hold step-wise. A positive squared response also orders the corresponding per-step norms, but monotonicity of the difference between raw norms requires additional scale conditions; Figure~\ref{fig:context-dose-response} is therefore empirical evidence consistent with the law rather than a direct proof of its assumptions.

\subsection{High-Probability Persistent Separation}

Theorem~\ref{thm:app-population-separation} derives a deterministic margin from Adapt-space geometry and contextual composition. We now quantify the reduction of stochastic EDM variation with $K$. Let $\mathcal H_{k}$ denote the trajectory filtration and define
\begin{equation}
\eta_i^{(k)}=r_i^{(k)}-\E[r_i^{(k)}\mid\mathcal H_{k-1}],
\qquad
\bar m_i=\frac{1}{K}\sum_{k=0}^{K-1}\E[r_i^{(k)}\mid\mathcal H_{k-1}].
\label{eq:app-martingale-noise}
\end{equation}

\begin{theorem}[Persistent-score concentration]
Suppose $\eta_i^{(k)}$ is conditionally $\sigma_i$-sub-Gaussian, i.e., for all $t\in\R$,
\begin{equation}
\E[\exp(t\eta_i^{(k)})\mid\mathcal H_{k-1}]
\leq\exp(t^2\sigma_i^2/2).
\label{eq:app-subgaussian-condition}
\end{equation}
Then
\begin{equation}
\Pr\!\left(|\mathrm{score}_i-\bar m_i|\geq t\right)
\leq2\exp\!\left(-\frac{Kt^2}{2\sigma_i^2}\right).
\label{eq:app-persistent-concentration}
\end{equation}
\end{theorem}

\begin{Proof}
Iterating the conditional moment-generating-function bound produces an exponential supermartingale for $\sum_k\eta_i^{(k)}$. Chernoff's method applied to both tails, followed by division by $K$, yields Eq.~\eqref{eq:app-persistent-concentration}.
\end{Proof}

Consequently, for a fixed anomaly--normal pair with conditional-mean gap $\bar m_A-\bar m_N$, both scores deviate by at most
\begin{equation}
t_\delta=\sigma\sqrt{\frac{2\log(4/\delta)}{K}},
\label{eq:app-pair-threshold}
\end{equation}
with probability at least $1-\delta$ when $\sigma_A,\sigma_N\leq\sigma$. The ordering is therefore preserved whenever $\bar m_A-\bar m_N>2t_\delta$. This formalizes the $K^{-1/2}$ variance reduction of persistent averaging under the stated martingale condition. Thus $K$ reduces stochastic variation, while its effect on the underlying mean gap remains governed by the dynamics.

For an all-node statement, if $\sigma_i\leq\sigma_T$ for every node, a union bound gives
\begin{equation}
\max_i|\mathrm{score}_i-\bar m_i|
\leq t_T:=\sigma_T\sqrt{\frac{2\log(2N/\delta_T)}{K}}
\label{eq:app-all-node-threshold}
\end{equation}
with probability at least $1-\delta_T$. Hence every detectable anomaly is ranked above every compatible normal node whenever their conditional-mean margin is uniformly larger than $2t_T$. The structural term in Eq.~\eqref{eq:app-population-margin}, the degree-dependent context error in Eq.~\eqref{eq:app-degree-context-bound}, and the temporal term in Eq.~\eqref{eq:app-all-node-threshold} respectively expose the roles of class separation, neighborhood sample size, and persistence. Their joint application rests on the stated prototype, conditional-independence, and martingale assumptions.

\section{Additional Method and Training Details}
\label{app:method-details}
\label{app:edm}

\subsection{Autoencoder objective}
The graph convolutional autoencoder reconstructs node attributes and graph structure with
\begin{equation}
\mathcal{L}_{\mathrm{AE}}=\frac{1}{N}\sum_{i=1}^{N}\lambda_x\|\vx_i-\widehat{\vx}_i\|_2
+\frac{1}{N}\sum_{i=1}^{N}\lambda_A\|\mA_{i,:}-\widehat{\mA}_{i,:}\|_2+\lambda_e\mathcal{L}_{\mathrm{edge}}.
\label{eq:app-ae-loss}
\end{equation}
For large graphs, the dense structure term is omitted to control memory. All nodes participate in this transductive objective, so the resulting initialization remains graph-conditioned.

This section documents the node-wise EDM implementation~\citep{karras2022elucidating} used by DDGAD. Training and inference operate on frozen AE latent embeddings. The MLP denoiser never receives graph adjacency; topology is introduced only by Combine.

\subsection{Forward perturbation.}
The forward process adds Gaussian noise with level $\sigma$ to an observed AE latent vector $\vx_0$:
\begin{equation}
\vx_\sigma = \vx_0 + \sigma\ve,
\qquad \ve \sim \mathcal N(\mathbf{0}, \mI).
\label{eq:app-edm-forward}
\end{equation}
The noise level $\sigma$ is sampled from a log-normal distribution with mean $-1.2$ and standard deviation $1.2$. We adopt the EDM parameterization as a generic continuous denoising framework and use fixed dataset-level noise ranges rather than borrowing the image-domain schedule directly.

\subsection{Training loss.}
The denoising network $D_\theta(\vx,\sigma)$ predicts the original observed latent $\vx_0$ from $\vx_\sigma$ by minimizing
\begin{equation}
\mathcal{L}_{\text{EDM}}
= \mathbb{E}_{\sigma,\vx_0,\ve}\!\left[
w(\sigma)\,
\bigl\|D_\theta(\vx_0+\sigma\ve,\sigma)-\vx_0\bigr\|_2^2
\right],
\label{eq:app-edm-loss}
\end{equation}
with the EDM weighting function
\begin{equation}
w(\sigma) = \frac{\sigma^2+\sigma_{\text{data}}^2}{(\sigma\cdot\sigma_{\text{data}})^2},
\qquad \sigma_{\text{data}} = 0.5.
\label{eq:app-edm-weight}
\end{equation}
This is the EDM preconditioning weight used by the implementation.

\subsection{Architecture and training.}
The denoiser is a node-wise MLP with a sinusoidal noise embedding, input projection, LayerNorm, SiLU activations, and an output projection back to the AE latent dimension. Its internal width is at least $128$. Training is full-graph, uses Adam, gradient clipping, a step learning-rate schedule, and early stopping on the unsupervised EDM training loss.

\subsection{Sampling.}
At each ATC iteration, DDGAD perturbs the current latent with fresh Gaussian noise and applies a power-law EDM schedule followed by Heun correction. Writing the numerical sampler as $\mathcal S_\theta$,
\begin{equation}
\psi_i^{(k)}=\mathcal S_\theta\!\left(\vz_i^{(k)}+\sigma_{\max}\boldsymbol{\xi}_i^{(k)};
\sigma_{\min},\sigma_{\max}\right),\qquad \boldsymbol{\xi}_i^{(k)}\sim\mathcal N(0,\mI).
\label{eq:app-edm-sampling}
\end{equation}
No edge index is passed to $\mathcal S_\theta$.

\subsection{Interpretation and boundary.}
EDM is trained on the full latent population rather than a label-filtered normal subset, so it provides a density-oriented node-wise reference rather than a guaranteed clean-manifold projector. At each step, conditional on the current graph-conditioned inputs $\vz^{(k)}$, Adapt introduces no new same-step cross-node aggregation; earlier Combine/Fuse updates may already have mixed neighboring information. Proposition~3 quantifies how local anomaly posterior mass biases the ideal denoising direction. Anomaly evidence arises only when this Adapt reference and the neighborhood context exhibit a persistent gap.

\section{Additional Experimental Details}
\label{app:experimental-details}

\subsection{Dataset statistics}
\begin{table}[H]
\centering
\caption{Dataset statistics.}
\label{tab:datasets}
\scriptsize
\begin{tabular}{lrrrr}
\toprule
\textbf{Dataset} & \textbf{Nodes} & \textbf{Edges} & \textbf{Features} & \textbf{Anomaly ratio} \\
\midrule
Books   & 1,418  & 3,695  & 21   & 1.97\% \\
Disney  & 124    & 335    & 28   & 4.84\% \\
Reddit  & 10,984 & 168,016 & 64  & 3.33\% \\
Amazon-5KN & 5,000 & 10,000 & 25 & 16.42\% \\
Amazon-Bal & 5,000 & 10,000 & 25 & 16.42\% \\
Tolokers & 11,758 & 530,758 & 10 & 21.82\% \\
\bottomrule
\end{tabular}
\end{table}

\subsection{Implementation and reporting protocol}
The autoencoder contains a two-layer GCN encoder and a two-layer GCN attribute decoder with a dot-product structure decoder, residual input projection, and layer normalization. We use Adam for both stages. The denoiser uses EDM preconditioning with $P_{\mathrm{mean}}=-1.2$, $P_{\mathrm{std}}=1.2$, $\sigma_{\mathrm{data}}=0.5$, and $\sigma_{\min}=0.002$. Early stopping monitors only the unsupervised EDM training loss.

Table~\ref{tab:inference-settings} records the Fuse weight $\alpha$ and ATC horizon $K$ used for the main benchmark results. Across all six benchmarks and five seeds, DDGAD uses a uniform default setting ($\alpha=0.7, K=6$) without dataset-specific tuning.
\begin{table}[H]
\centering
\caption{DDGAD inference settings used in Table~\ref{tab:main}.}
\label{tab:inference-settings}
\scriptsize
\begin{tabular}{lcc}
\toprule
\textbf{Dataset} & \textbf{$\alpha$} & \textbf{$K$} \\
\midrule
Books      & 0.7 & 6 \\
Disney     & 0.7 & 6 \\
Reddit     & 0.7 & 6 \\
Amazon-5KN & 0.7 & 6 \\
Amazon-Bal & 0.7 & 6 \\
Tolokers   & 0.7 & 6 \\
\bottomrule
\end{tabular}
\end{table}

\paragraph{Transductive evaluation partitions.}
The detector is trained transductively on the full graph: features and edges of every node, including nodes later assigned to the test partition, are available to the AE and EDM, whereas anomaly labels are never provided to either model. Labels are used only outside the detector to construct deterministic class-stratified evaluation partitions and to compute metrics. For each seed, 50\% of anomalous nodes and 20\% of normal nodes are first assigned to a train/validation pool; 20\% of each class-specific pool is then assigned to validation, and the remainder of that pool is assigned to train. All nodes not selected into the pool form the test evaluation partition. Up to integer rounding, the resulting train/validation/test proportions are therefore 40\%/10\%/50\% for anomalous nodes and 16\%/4\%/80\% for normal nodes. The names ``train'' and ``validation'' are retained for partition bookkeeping and compatibility across experiments: DDGAD does not use these masks for parameter optimization, early stopping, checkpoint selection, inference, or score construction. It produces one anomaly score for every node before the test mask is applied, and the reported benchmark AUROC is computed only on the fixed test partition.

Main-table values are the mean and sample standard deviation over five independently run seeds. Every method uses the same AUROC computation on its deterministic seed-specific test partition. Within each controlled ablation table, all rows share one checkpoint and evaluation protocol; independently trained ablation suites are interpreted through their within-table relative differences.

\paragraph{Main-table reruns.}
We rerun all main-table methods on the six stored graph instances with five seeds and report the resulting mean and sample standard deviation. GCN-AE is our label-free reconstruction control: a two-layer GCN is trained for 100 epochs with learning rate $0.01$ to reconstruct node attributes under mean-squared error, and the per-node squared reconstruction error is used as its anomaly score. GraphMAE is originally a self-supervised representation learner rather than a dedicated anomaly detector. We use the PyG implementation with 200 pretraining epochs, row-normalized features, a 0.5 mask rate, and the scaled-cosine-error objective with exponent 2; its node score is the mean masked-node reconstruction error over 20 independently sampled masks. SCAN is evaluated as a non-neural structural clustering baseline. The remaining baselines use their corresponding implementations and the seed-specific test partitions described above.

\subsection{Context-source intervention protocol}
\label{app:context-intervention}
The intervention in Section~\ref{sec:context-intervention} is a post-hoc mechanism analysis, not part of detector training or benchmark evaluation. For each Amazon graph, we load one pre-existing checkpoint, compute $\vz^{(0)}$ on the original graph once, and generate five stochastic realizations of the clean EDM trajectory. These node-reference snapshots are frozen before any intervention. Intervention edges are visible only to a uniform Combine operator; neither the GCN encoder nor EDM receives the modified adjacency.

Amazon-5KN and Amazon-Bal each contain 5,000 nodes, including 821 labeled anomalies. Among their 4,179 normal nodes, 1,454 and 1,395 nodes, respectively, are isolated under the adjacency used by Combine. We exclude these degree-zero nodes from the matched intervention because adding their first neighbor changes the existence of the contextual reference itself. This leaves 2,725 and 2,784 eligible normal victims. Each of 30 trials samples 500 victims and, for every victim, nested sets of $r\in\{1,2,4,8,16\}$ distinct non-neighbor sources. The anomaly-add and normal-add counterfactuals use the same victim and edge count; sources differ only in post-hoc label-defined type.

Uncertainty is computed over the 30 trial-level paired mean effects. Figure~\ref{fig:context-dose-response} uses 50,000-resample percentile bootstrap confidence intervals at each dose. Table~\ref{tab:context-intervention} reports the same interval for persistent EDM at $r=16$ and an exact two-sided Wilcoxon signed-rank test of the 30 trial-level paired effects against zero. The standardized effect divides the mean paired source effect by the clean-score standard deviation among normal nodes. ``Victim $+$'' averages, over trials, the within-trial fraction of victims satisfying $s_{\mathrm{anomaly\text{-}add}}>s_{\mathrm{normal\text{-}add}}$.

Including isolated victims changes the estimand by simultaneously introducing the first contextual reference and can reverse the aggregate response; we therefore restrict the source-controlled comparison to non-isolated victims and the equal-degree normal-source counterfactual.

\end{document}